\documentclass[10pt,amssymb,twocolumn,notitlepage]{article}

\usepackage{epigraph}
\usepackage[affil-it]{authblk}
\usepackage{dcolumn}
\usepackage[stable]{footmisc}
\usepackage{blkarray}
\usepackage{bm}
\usepackage[mathscr]{euscript}
\usepackage[font=scriptsize]{caption}
\usepackage{subcaption,graphicx}
\usepackage[table]{xcolor}
\usepackage[margin=1.7cm]{geometry}
\usepackage{tcolorbox}
\usepackage{hyperref}
\usepackage{gensymb}
\usepackage[font=footnotesize]{caption}
\usepackage{hyperref}
\hypersetup{
    colorlinks=true,
    linkcolor=blue,
    filecolor=magenta,      
    urlcolor=blue,
}
\usepackage{steinmetz}
\usepackage{amssymb}
\usepackage{makecell}
\usepackage{algorithm,lipsum,xcolor,caption}
\usepackage{enumitem}
\usepackage{amsmath,esint}
\usepackage{fancyvrb}
\usepackage{xcolor}
\usepackage{arydshln}
\usepackage{mathtools}

\definecolor{newblue}{rgb}{0.0, 0.28, 0.67}
\definecolor{newgreen}{rgb}{0.13, 0.55, 0.13}
\definecolor{newred}{rgb}{0.87, 0.72, 0.53}

\newcommand{\R}{\mathbb{R}}

\definecolor{newblue}{rgb}{0.0, 0.28, 0.67}
\definecolor{newgreen}{rgb}{0.13, 0.55, 0.13}
\definecolor{newred}{rgb}{0.87, 0.72, 0.53}
\usepackage{stackengine}

\title{Further Generalizations of the Jaccard Index}
\author{Luciano da Fontoura Costa \\ \emph{luciano@ifsc.usp.br}}
\affil{S\~ao Carlos Institute of Physics -- DFCM/USP} 
\date{10th Oct 2021}

\begin{document}

\twocolumn[
\begin{@twocolumnfalse}
    \maketitle
    \begin{abstract}
Quantifying the similarity between two mathematical structures or datasets constitutes a  particularly interesting and useful operation in several theoretical and applied problems.  Aimed at  this specific objective, the Jaccard index has been extensively used in the most diverse types of problems, also motivating some respective generalizations.   The present work addresses further generalizations of this index, including its modification  into a coincidence index capable of accounting also for the level of relative interiority between the two compared entities, as well as respective extensions for  sets in continuous vector spaces,  the generalization to multiset addition, densities and  generic scalar fields, as well as a means to quantify the joint interdependence between  two random variables.  The also interesting possibility to take into account more than two  sets has also been addressed, including the  description of an index capable of quantifying  the level of chaining between three structures.   Several of the described and suggested  eneralizations have been illustrated with respect to numeric case examples.  It is also posited that these  indices can play an important role while analyzing and integrating datasets in modeling approaches and pattern recognition activities, including as a measurement of clusters  similarity or separation and as a resource for representing and analyzing complex networks.
     \end{abstract}
\end{@twocolumnfalse} \bigskip
]

\setlength{\epigraphwidth}{.49\textwidth}
\epigraph{`\emph{Riedificano Ersilia altrove. Tessono com i fili 
una figura simile che vorrebbero pi\'u complicata e insieme pi\'u regolare dell'altra.}'}
{\emph{Italo Calvino, Le Citt\`a Invisibili.}}

\section{Introduction}

Despite its seeming simplicity, set theory underlies a substantial portion of the 
mathematical and physical sciences, while being also extensively used in virtually every 
area of human activity.

In fact, set theory concepts has been so ubiquitous as to have been incorporated into 
common language and daily conversations.  When one says ``I will buy bananas and 
potatoes and tomatoes,''  it is actually the set operation of union that it is being meant.  
Interestingly, the tenuous border between set theory and propositional logic is often 
blurred by humans (see~\cite{CostaAmple,CostaAnalogies}).  At the same time, multisets 
(e.g.~\cite{Samanthula,Bacciu,Hein,Knuth,Blizard,Blizard2,Thangavelu,Singh,CostaMset})
offer means for extending set so as to consider  the multiplicity of elements which, in 
a sense, seems at least as compatible with human intuition than the classic set theory.

Other concepts that are as ubiquitously employed in every human activity regards the
similarity and distance between two entities.  Mathematically, this can
be related to quantifying in an objective manner several types of similarity between
two or more mathematical structures such as scalars, sets, vectors, matrices,
functions, densities, graphs, etc (e.g.~\cite{Kavitha,Mirkin,Jarvis,Quintana,Wolda}).   
This can be done in several manners, which frequently
take into account the respective type of structure.  For instance, vectors are often
compared in terms of their inner product, and several similarity indices 
(e.g.~\cite{Steinley}) have been suggested for comparing matrices with binary features.

One approach to the similarity between two sets that has attracted particular attention as a 
consequence of its simple and intuitive conceptual characteristics, being therefore employed
extensively, is the \emph{Jaccard} or \emph{Tanimoto} index (e.g.~\cite{Jac:wiki,Jaccard1,BioJaccard,Tanimoto}).
In addition to being constrained within the interval $[0,1]$, the Jaccard index
requires little computational expenses.   Besides its vast range of applications (e.g.~\cite{Yuan,Jac:wiki,Lieve,Loet,Park}), most of them related to binary or
categorical data, the Jaccard index has also motivated some
extensions and generalizations, including its adaptation to discrete multisets 
with positive multiplicites (e.g.~\cite{Samanthula,Bacciu,CostaMset}).

Given the popularity of the Jaccard index, as well as its appealing characteristics,
it would be particularly useful if it could be adapted to as many as possible other
mathematical structures. The present work aims at developing further possible 
generalizations of the Jaccard index.

We start by providing a brief historic perspective of the Jaccard index, which was
introduced in 1901 by Paul Jaccard (1868--1944) under the name of
\emph{coefficient de communaut\'e}.

Subsequently, we focus on the relative limitation of this index to reflect to which level one
set is contained into the other, and a respective adaptation of the Jaccard index is
then proposed to address this limitation that involves another measurement between
two sets, here called \emph{interiority index} (also known as overlap index e.g.~\cite{Kavitha}).  
More specifically, we define the  \emph{coincidence index} between two sets as 
corresponding to the square root of the product of  respective Jaccard and Interiority indices.
The specific characteristics and properties of these two indices, as well as some additional
alternatives, are illustrated through a graphic construction, and it is shown that the 
coincidence index impose a more strict and selective characterization of the similarity
between two sets.   Another variation of the Jaccard index, here called addition-based
Jaccard similarity, is then also motivated and presented.
 
We then approach the adaptation of the Jaccard index to take into account sets
corresponding to regions in continuous spaces such as $\R^N$.  It is shown that this can be immediately
accommodated into the standard Jaccard  (and also the coincidence) indices by having
the regions area in place of the sizes of sets.  In addition to allowing useful graphical
characterizations of the Jaccard and coincidence indices, this extension to continuous
sets also paves the way to dealing with densities and scalar fields.  In particular, we
develop a related graphical construct to compare the relationship between the Jaccard,
interiority and coincidence indices.

Next, we address the particularly interesting question of adapting the Jaccard index
to become capable of comparing \emph{densities},  \emph{functions} and
\emph{fields} in continuous spaces $\R^N$.   This is achieved by extending multisets 
to real function spaces~\cite{CostaMset,CostaAnalogies}, allowing a version of the 
Jaccard and coincidence indices incorporating integrals (functionals) of the minimum 
(multiset intersection) and maximum (multiset union) operations
along the respective space. Because generic fields involve negative values, the
approach is presented first for non-negative densities, being subsequently extended
to more general fields with negative multiplicities.  The potential of the approach,
which is conceptually and computationally simple, is then illustrated with respect to
comparing probability density functions as well as more generic functions corresponding
to two sinusoidals as well as two real-world images.

Another issue of special relevance that has been addressed regards the inherent, but not
often considered, relationship between the quantification of the similarity between two 
probability densities with the also ample subject of characterizing joint variation of two 
random variables.   Two particular problems are addressed.  We show that the multiset 
Jaccard adaptation to densities and functions can be effectively
applied to quantify the joint relationship between two random variables, be it in terms
of discrete observations or while taking into account their probability densities describing
standardized versions of the involved variables.  

The next generalizing approach described in the present work concerns the possibility of generalizing the
Jaccard index to deal with more than 2 sets (see also~\cite{Quintana}).  
We argue that there are two main ways
in which this problem can be addressed.  First, it is possible to have any of the two
sets involved in the Jaccard index to correspond to generic combinations of any number
of sets, obtained by using set operations.  Alternatively, more than two sets can be
actually considered as arguments of extended Jaccard indices.  The latter possibility
has been illustrated through the development of a generalization of the Jaccard index
capable of quantifying the degree of chaining between three sets, as intermediated
by one of them.

The above developments motivate the possibility to systematically combine several normalized
indices, possibly involving several orders of data combinations, so as to logically
integrate diverse data characteristics of interest about the analyzed data, therefore leading to
a possible algebra of indices, of which the coincidence index constitutes an example.

The article concludes by discussing the particularly important role of indices such as
those discussed and suggested here for the ubiquitous activities of model building
and pattern recognition.  Prospects for future developments are also provided.

\section{A Brief Historic Note on Paul Jaccard}

Paul Jaccard (1868--1944) (e.g.~\cite{Jac:wiki,BioJaccard,BioJaccard2}) was a researcher in the area 
of plant physiology in Zurich, who
started his studies in 1889 at the L\ Universit\'e de Lausanne (paleobotanic and
phytoembriology), then moving to the 
L'Universit'e de Zurich, where he concluded his PhD in 1894, followed by an
internship in Paris with  Gaston Bonnier (1853--1922), a French botanist and ecologist
and full professor at Sorbonne.

Jaccard held teaching and research activities at Lausanne, and then Zurich, 
focusing on subjects related to geobotanic and tree histophysiology, including wood 
microscopic studies.  He also travelled 
extensively to Egypt, Sweden and Turkistan (Kazakhstan region), being particularly interested
on trees interbreeding from the anatomic and physiologic points of view.   

The similarity index that bears his name was proposed in 1901~\cite{Jaccard1} as a means to quantify
co-localization of alpine flora, with particular interest in the study of species diversity.
The index that now bears his name can be expressed in set theory notation as:
\begin{equation}
   J(A,B) = \frac{|A \cap B|}{|A \cup B|}
\end{equation}

where $A$ and $B$ are any two sets and $|A|$ and $|B|$ are their respective
cardinality (number of elements).

Jaccard also proposed another relative index~\cite{Jaccard2}, namely the 
\emph{coefficient g\'en\'erique}, aimed at quantifying species-to-genus ratio, which consists of:
\begin{equation}
  G(species,genus) = \frac{|genus|}{|species|}
\end{equation}

where $|genus|$ and $|species|$ stand respectively to the number of genus and species
in a considered region.

\section{The Basic Jaccard Index}

The basic Jaccard index can be simply expressed (e.g.~\cite{Jaccard1,Jac:wiki,BioJaccard}) as:
\begin{equation}  \label{eq:Jac}
   \mathcal{J}(A,B) =   \frac{\left| A \cap B \right|}  {\left| A \cup B \right|} =
     \frac{\left| A \cap B \right|}  {\left| A \right| + \left| B \right| - \left| A \cap B \right|}
\end{equation}

where $A$ and $B$ are any two sets to be compared.

It is interesting to keep in mind that, though not frequently specified, the universe
set of $A$ and $B$ can be conveniently taken as being equal to $\Omega = A \cup B$.

The Jaccard distance can be immediately derived from the Jaccard index by making:
\begin{equation}
   \mathcal{D}_J(A,B) = 1 - \mathcal{J}(A,B) 
\end{equation}

This approach can be immediately extended to any other
similarity index bound between 0 and 1.

It is also interesting to observe that it is possible to modify the Jaccard index so as to reflect 
in absolute terms the effective cardinality of the intersection of the two sets.  This can be done as:
\begin{equation}  \label{eq:Jac}
   \mathcal{J}_2(A,B) =   \frac{\left| A \cap B \right|^2}  {\left| A \cup B \right|} 
\end{equation}

with $0 \leq J_2(A,B) \leq |A \cap B|$.

Actually, it is also possible to consider taking higher powers, therefore implying even
larger intersection cardinality weights, such as:
\begin{equation}  \label{eq:Jac}
   \mathcal{J}_P(A,B) =   \frac{\left| A \cap B \right|^P}  {\left| A \cup B \right|} 
\end{equation}

with $P \in \R$.  

The Jaccard index can be immediately generalized to \emph{multisets} or \emph{bags} (e.g.~\cite{Hein,Knuth,Samanthula}),  which are  basically sets in which repeated elements are allowed.   

The multisets $A$ and $B$ sharing the same elements (support) can be simply represented 
as respective vectors $\vec{A} = [a_1, a_2, \ldots, a_N]$, $\vec{B} = [b_1, b_2, \ldots, b_N]$, 
where $N$ is the total number of possible distinct elements in the universe defined
by the union of the two multiset elements, and $a_i$ corresponds to the multiplicity of 
element $i$ in the multiset $A$.  The Jaccard index for multisets then becomes:
\begin{equation}  \label{eq:Jac_multi}
   \mathcal{J}_M(A,B) =   \frac{\sum_{i=1}^N\min{(a_i,b_i)}}  {\sum_{i=1}^N\max{(a_i,b_i)}} 
 \end{equation}
 
with $0 \leq \mathcal{J}_M(A,B) \leq 1$.
 
As an example, let's consider $A = \left\{ a, a, a, b, b \right\}$ and $B = \left\{ a, a, b, c, c, d \right\}$.
If we have the set of possible elements organized into the indexing vector $\vec{p} = [a, b, c, d]$, 
we will obtain $\vec{A} = [3, 2, 0, 0]$ and $\vec{B} = [2, 1, 2, 1]$.  Observe that the order of elements 
in $\vec{p}$ is immaterial to our analysis.  The, we have:
\begin{equation}  \label{eq:Jac_multi}
   \mathcal{J}(A,B) =   \frac{2 + 1 + 0 + 0}  {3 + 2 + 2 + 1} = \frac{3} {8} 
 \end{equation}
 
As a consequence, this adaptation of the Jaccard index allows it to be applied also to vectors,
matrices, and graphs.  In the case of matrices, the Jaccard equation
can be further modified as:
\begin{equation}  \label{eq:Jac_matrix}
   \mathcal{J}_M(A,B) =   \frac{\sum_{i=1}^N \sum_{j=1}^N\min{(a_{i,j,}b_{i,j})}}  
   {\sum_{i=1}^N\sum_{j=1}^N \max{(a_{i,j},b_{i,j})}} 
 \end{equation}
 
Observe that many other mathematical structures, such as matroids, tensors, etc., 
can be compared by further adapting the above equation.

\section{Interiority and Coincidence Indices}

\begin{figure}[h!]  
\begin{center}
   \includegraphics[width=0.7\linewidth]{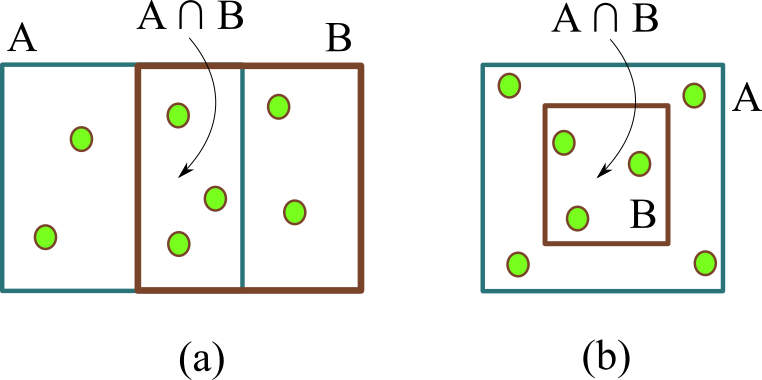}  
    \caption{Two distinct situations involving two sets $A$ and $B$ that yield
    the same Jaccard index value of $3/7$.   However, the two sets in (b)
    are much more compatible because $B$ is a subset of $A$ and therefore
    shares all its elements. }
    \label{fig:motiv}
    \end{center}
\end{figure}

As illustrated in the previous sections, and also by the relatively extensive related literature,
the Jaccard index provides an intuitive and logical manner to quantify the similarity
between two discrete or continuous set.  Yet, there is one particular situation, illustrated
in Figure~\ref{fig:motiv}, which is not accounted for by this index.

\vspace{0.5cm}

As it can be easily verified, both the situations depicted in Figure~\ref{fig:motiv} lead to the
same Jaccard index $\mathcal{J} = 3/7$.  However, the situation in (b) can be
deemed to be quite distinct because,
in this case, the set $B$ is completely contained in $A$ to the point of becoming a subset,
i.e.~$B \subset A$.  In other words, all elements of $B$ are shared with the set $A$.
This is not the case in the situation (a), for both sets $A$ and $B$ have elements that are
not shared. 

It therefore follows that it would be interesting to obtain a modification of the Jaccard index
that could distinguish between these two situations.  A possible approach is described as
follows.

We start by considering an index capable of quantifying how much a set is relatively interior to another.
Let $A$ and $B$ be any two sets.  The henceforth called \emph{interiority index} (also overlap
index e.g.~\cite{Kavitha}) can be written as:
\begin{equation} \label{eq:Int}
   \mathcal{I}(A,B) =   \frac{\left| A \cap B \right|}  {\min \left\{ \left| A \right|, \left| B \right| \right\} }
\end{equation}

Though this index has been also known by the names of overlap or homogeneity, in this
work we will adhere to the \emph{interiority} term as it seems to convey more directly the
concept of how much one of the sets is contained in the other.

It can be verified that $0 \leq \mathcal{I}(A,B) \leq 1$.  Its minimum value is observed when
$A$ is completely separated from $B$, i.e.~$A \cap B = 0$.  The maximum value is reached
when any of the sets is completely contained into the other.  In other words, there is no
need to specify which of the two sets is being considered as being internal to the other.

By comparing Equations~\ref{eq:Jac} and~\ref{eq:Int}, it follows that:
\begin{equation} 
   0 \leq \mathcal{I}(A,B)  \leq \mathcal{J}(A,B)  \leq 1
\end{equation}

and it can also be verified that:
\begin{equation}
   0 \leq  \mathcal{J}(A,B) \; \mathcal{H}(A,B)  \leq 1
\end{equation}

The verification of similarity accounted for by the Jaccard index can be conveniently
combined with the interiority index simply by considering their respective product,
i.e.:
\begin{equation}
    \mathcal{C}(A,B) = \mathcal{J}(A,B) \; \mathcal{I}(A,B) 
\end{equation}

which is the same as:
\begin{equation}
    \mathcal{C}(A,B) =  \frac{\left| A \cap B \right|^2}  {\left| A \cup B \right| \; \min \left\{ \left| A \right|, \left| B \right| \right\} }
\end{equation}

It may also interesting to take the square root of the coincidence in order to
compensate for the  smaller values implied when multiplying to measurements
in the interval $[0,1]$.  Therefore, in the present work we will adopted the
coincidence indes as:
\begin{equation}
    \mathcal{C}(A,B) = \sqrt{ \mathcal{J}(A,B) \; \mathcal{I}(A,B) }
\end{equation}

It is also possible, in certain situations, to dispense with the square root operation.

In some specific cases it is also possible to use these two indices separately, 
defining a corresponding tuple $[\mathcal{J}(A,B),\mathcal{I}(A,B)]$.

As with the Jaccard index, the coincidence index can also be generalized to virtually
any mathematical structure including, we will discuss in Section~\ref{sec:func} functions and
fields in $\R^N$.

\section{Weighted Discrete Elements}

The Jaccard and coincidence indices can be readily adapted (e.g.~\cite{Jac:wiki})
to cope with cases in which the
elements of sets $A$ and $B$ have been assigned respective weights corresponding to their
relative importance in each specific problem.   

This situation can be approached by using 
ordered pairs to represent each of the elements in $A \cup B$ associated to its respective 
weigh, i.e.~$[x_i, w(x_i)]$.    The Jaccard index then becomes:
\begin{equation}  \label{eq:Jac}
   \mathcal{J}_W(A,B) =   \frac{   \sum_{x_i \in (A \cap B)} w(x_i) }  { \sum_{y_i \in (A \cup B)} w(y_i) } 
 \end{equation}

with $0 \leq \mathcal{J}_w(A,B) \leq 1$.

As an example, let $A = \left\{ [a,2]; [b, 5]; [c,1] \right\}$ and $B = \left\{ [b, 5]; [e,1], [f,1] \right\}$.
It follows that $A \cap B = \left\{  [b, 5] \right\}$.
\begin{equation}  \label{eq:Jac}
   \mathcal{J}_W(A,B) =   \frac{ 5 }  { 10 } = \frac{1}{2}  
 \end{equation}
 
Thus, in spite of in this particular example the intersection being limited to a single of the 
possible elements, the Jaccard
index resulted relatively high as a consequence of the large weigh associated to the element $b$.

Observe that the weighted version of the Jaccard index is not the same as the Jaccard index
adapted to multisets, as the latter case does not involve the sum of weights.  However, it
is possible to considered weighted multisets, in which case the Jaccard index becomes:
\begin{equation}  \label{eq:Jac}
   \mathcal{J}_{[W,M]}(A,B) =   \frac{   \sum_i{w_i \min \left\{a_i, b_i \right\}}}  { \sum_i{ w_i\max \left\{a_i, b_i \right\} } } 
 \end{equation}

\section{Addition-Based Multiset Jaccard Index}
  
The multiset Jaccard index can be further generalized by taking into account the
\emph{sum} of the two sets $A$ and $B$ allowed by multiset theory,
instead of their respective union, which leads to:
\begin{equation}
  \mathcal{J}_S(A,B) =  \frac{2 \sum_{i=1}^N\min{(a_i,b_i)}}  {\sum_{i=1}^N \left( a_i+b_i \right)}
\end{equation}

with $0 \leq \mathcal{J}_S(A,B) \leq 1$.  Observe that other multiset operations including
subtraction, complement, and intersection can also be employed to define other indices.  

The interesting feature of this index is that it takes into account the situations where
the multiple instances in the multisets need to be taken into account at its fullest
when combining the sets.   

As an example, let's consider that $A = \left\{a, a, a, b, c, c, c \right\}$ and
$B = \left\{ a, a, c \right\}$.  Then, we have that:
\begin{equation}
  \mathcal{J}_S(A,A) = \frac{(2)(7)}   {7+7} = 1
\end{equation}

and:
\begin{equation}
  \mathcal{J}_S(A,B) = \frac{(2)(3)}   {7 + 3} = \frac{3}{5}
\end{equation}

The additive multised Jaccard index can be immediately combined with
the respective interiority index to yield the \emph{addition-based multiset coincidence
index}.

\section{Continuous Sets}

The henceforth described approach holds for $\R^N$, but we shall
consider the plane vector space $\R^2$.  It is possible to associate sets to the
points $(x,y)$ of this space in any possible manner, such as 
$R = \left\{ (x,y) \; |\; x \text{ and } y \text{ are even} \right\}$, which is a discontinuous in $\R^2$,
or $S = \left\{ (x,y) \; |\; 0 \leq x \leq 2, -1 \leq y \leq 1 \right\}$, which defines a continuous
region.  

\begin{figure}[h!]  
\begin{center}
   \includegraphics[width=0.9\linewidth]{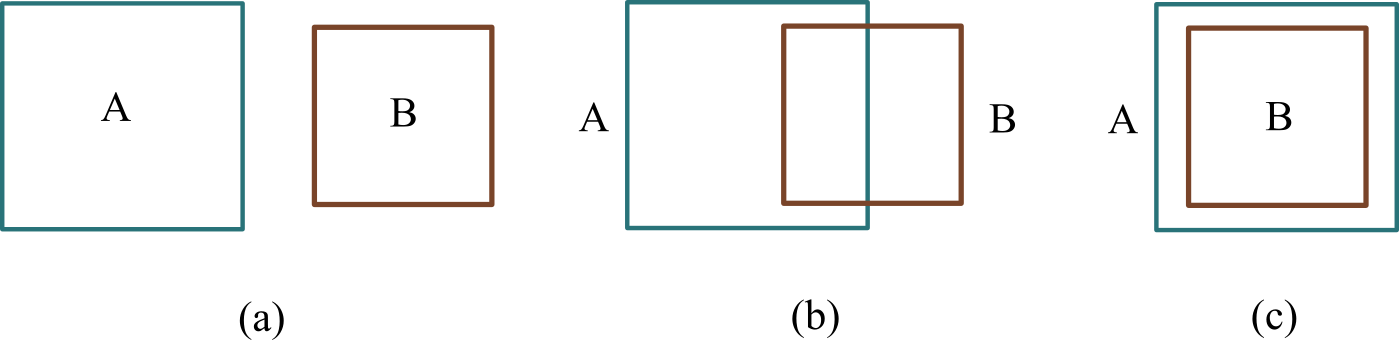}  
    \caption{The three most relevant situations to be considered when comparing two sets:
    (a) no intersection; (b) partial interesection; (c) complete intersection.}
    \label{fig:plane}
    \end{center}
\end{figure}

Though the Jaccard index can be immediately applied to any of these sets, it is
of particular interest to our developments to consider sets configurations
corresponding to simple connected regions of $\R^2$ such as those
illustrated in Figure~\ref{fig:plane}.

\vspace{0.5cm}

In this case, the size of the involved sets and subsets can be conveniently represented by 
the respective areas, 
indicated as $|A|$, $|B|$, $|A \cap B|$, and $|A \cup B|$, which can be immediately 
used in Equation~\ref{eq:Jac}.

The three cases in Figure~\ref{fig:plane} correspond to the most representative situations 
when comparing two sets.
In Figure~\ref{fig:plane}(a), we have two separated sets, which results in null intersection,
suggesting minimal similarity between the two sets. The situation depicted in (c) can be understood 
as leading to the maximum similarity that can be achieved with the sets $A$ and $B$.  
Figure~\ref{fig:plane}(b) illustrates a frequently found situation in which there is some intersection
between the sets.  In this case, the similarity value would be expected to increase with the
intersection area in a possibly linear manner.

\begin{figure}[h!]  
\begin{center}
   \includegraphics[width=0.5\linewidth]{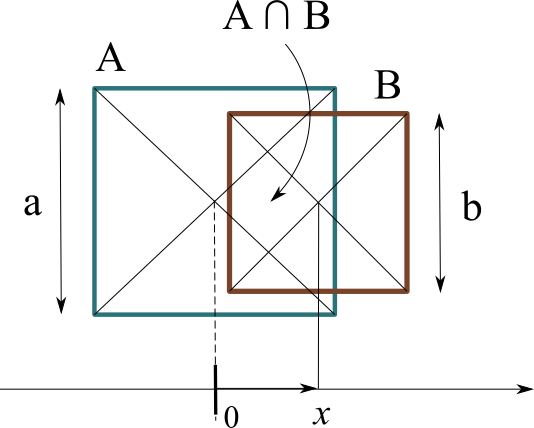}  
    \caption{A construction representing all possible situations regarding the similarity of two
    sliding squares $A$ and $B$ with sides $a$ and $b$, respectively. 
    Without loss of generality, we assume that $a \geq b$.  Any of these situations
    can be specified by just two parameters: the relative position $x$ and the relative size
    $r = b/a$.  This construction allows us to better understand the behavior of the Jaccard
    and other similarity indices covered in this work.}
    \label{fig:constr}
    \end{center}
\end{figure}

The situation represented in Figure~\ref{fig:plane}(b) actually incorporates the two other 
situations as limit cases.  Consider the diagram shown in Figure~\ref{fig:constr},
involving two square regions $A$ and $B$, with respective sides $a$ and $b$, 
$b \leq a$.

\vspace{0.5cm}

The relative position, and also the similarity, of the two sets can be completely controlled in terms 
of the relative position parameter $x$, with $\frac{a-b}{2}  \leq x \leq \frac{a+b}{2}$.
As $b$ increases, the two squares progressively separate, therefore becoming less similar.

There is only one other parameter that needs to be specified in order to completely represent
the situation in  Figure~\ref{fig:constr}, namely the relative sizes of the two regions $r = b/a$,
with $0 \leq r \leq 1$.

The area of the intersection $A \cap B$ and union $A \cup B$ of the two sets can now be
conveniently expressed in terms of $x$ and $b$ as:
\begin{eqnarray}
|A \cap B| = (r a) \left(\frac{a}{2} - \left(x- \frac{r a}{2} \right) \right) = \nonumber \\
  \frac{1}{2} \left(a^2 r (1+r) - 2 r a x \right)   \\
|A \cup B| = \frac{1}{2} \left( 2 a^2 \left(1 + r^2 \right)  -a^2 r (1+r) + 2 r a x \right)
\end{eqnarray}

We can now rewrite the Jaccard index as:
\begin{equation}  \label{eq:Jac}
   \mathcal{J}(A,B) =   \frac{a^2 r (1+r) - 2 r a x }  { 2 a^2 \left(1 + r^2 \right)  -a^2 r (1+r) + 2 r a x  } 
\end{equation}

\begin{figure*}[h!]  
\begin{center}
   \includegraphics[width=0.9\linewidth]{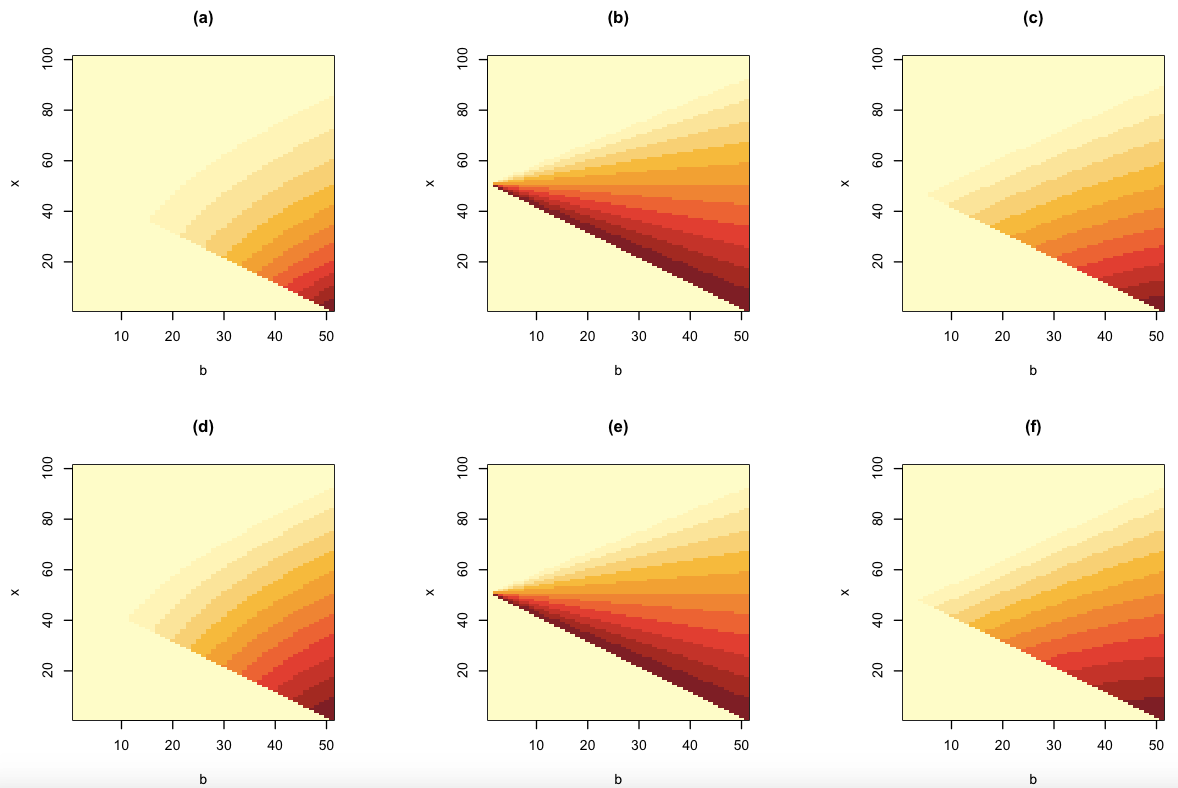}   \\
    \caption{The Jaccard (a), interiority (b), and square root of the coincidence (c) indices
    obtained for the geometrical construction illustrated in Figure~\ref{fig:constr}. 
    The heat map increases from yellow to brown.  The incorporation of the
    interiority level into the Jaccard index leads to a more comensurated
    distribution of the level sets. The maximum value the Jaccard and
    continuity indices are to be found at the lower righthand corner of the
    respective plots.  The values of the interiority index increase linearly
    from the top to the bottom diagonal in (b).  Along (d) to (f), we have
    analogous results concerning the additive multiset Jaccard index (d), the 
    interiority (e), and square root of the additive coincidence index,
    the latter corresponding to the p (f). The latter index corresponds to
    the product of the additive multiset Jaccard and interiority indices.}
    \label{fig:Jac_surf}
    \end{center}
\end{figure*}

Figures~\ref{fig:Jac_surf} (a) to (c) present the Jaccard index for two rectangles, as developed above,
in terms of several configurations of the parameters $x$ and $r$.
Observe the yellow regions obtained for all indices, corresponing to the situations
in which the smaller square is either completely insider or outside the other square.

The first row in Figure~\ref{fig:Jac_surf} concerns the Jaccard (a), interiority (b) and 
respectively obtained coincidence (c) indices.  It can be observed that the Jaccard index
values increase diagonally from top right to the lower left corner, achieving the peak
$a$ at the bottom left corner in (a).  The interiority index, however, varies from 0 to 
1 along each vertical slice in (b), being therefore `degenerate' in the mathematical
sense of not implementing a bijective mapping with respect to the peak 1.  Indeed, every point
at the lower non-null region in (b) will correspond to maximum interiority of $1$.
The coincidence index values, shown in (c) correspond to the product between the
Jaccard and interiority indices, yielding a bijective association with the maximum value 
of 1 at the bottom right corner.  Observe the respective change in the level sets
shapes.  Similar results hold for the addition-based Jaccard index shown in the
lower row in Figure~\ref{fig:Jac_surf}(d-f).

\vspace{0.5cm}

\begin{figure*}[h!]  
\begin{center}
   \includegraphics[width=1\linewidth]{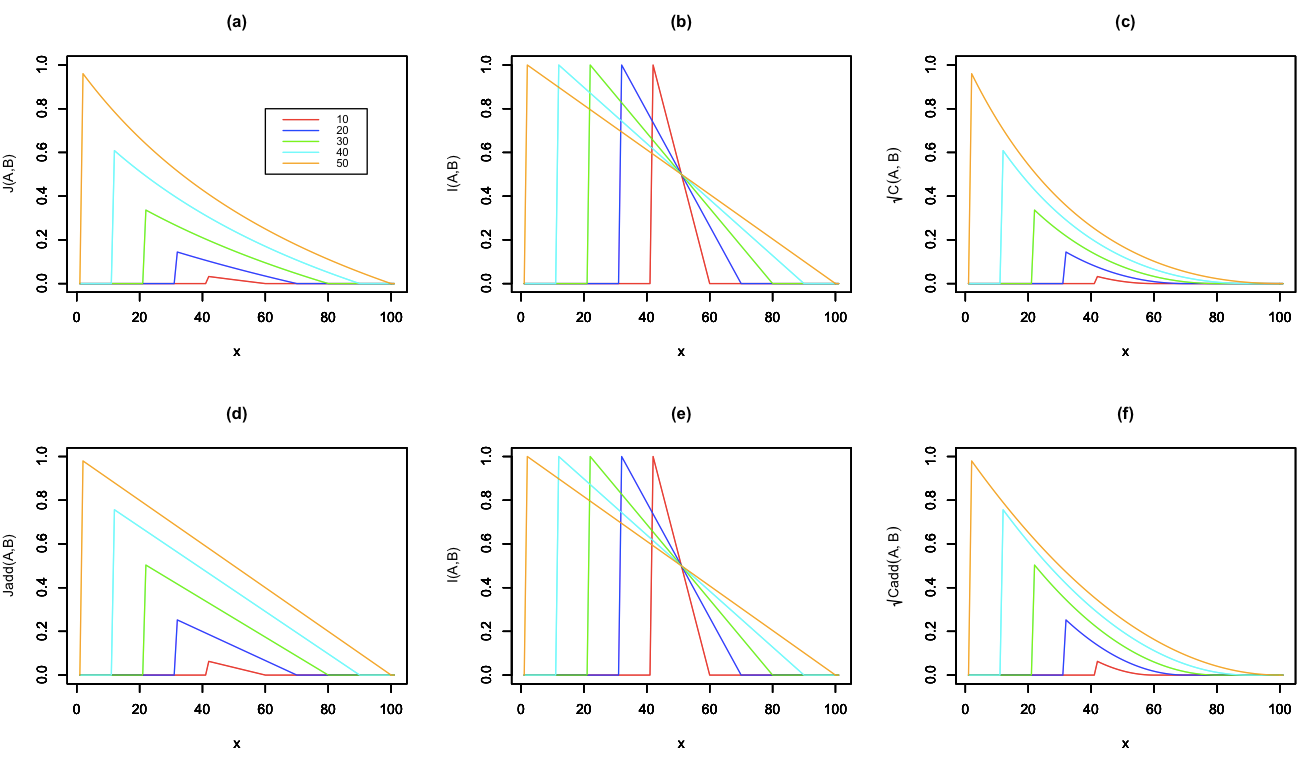}   \\
    \caption{Five vertical slices of the respective indices in Fig.~\ref{fig:Jac_surf}.  Only the interiority
    and additive multiset Jaccard indices account for the expected linear decrease of
    similarity following $x$ displacements.  However, the interiority is unable to take into
    account the relative size of the squares, which leaves only the additive multiset Jaccard
    as presenting a complete and linear quantification of the similarity between the two sets.
    The square root of the coincidence indices both penalizes the similarity when $x$ is small, 
    with the basic coindicence index imposing the most severe test of similarity.   }
    \label{fig:slices}
    \end{center}
\end{figure*}

The geometrical construct in Figure~\ref{fig:constr} therefore provides an interesting 
approach to comparing the varying results obtained in Figure~\ref{fig:Jac_surf}.  The rationale is as 
follows:  as the square $B$ slides from being completely inside the square $A$, until
it becomes completely outside the latter, it is reasonable to expect the
similarity to decrease in a linear manner with $x$.  This suggests that we can compare the several
indices in  Figure~\ref{fig:Jac_surf} while taking into account their respective slice along 
vertical slices of the scalar fields.   For simplicity's sake, we will consider five slices 
corresponding to $b = 10, 20, 30, 40, 50$.  The results are shown in Figure~\ref{fig:slices}.

\vspace{0.5cm}

Among the five indices in Figure~\ref{fig:Jac_surf}, only the additive multiset Jaccard index
accounts for linear similarity quantification as $x$ varies also in linear manner.  The
interiority, as expected, is unable to consider the relative size of the sets.  The coincidence
indices penalize the similarity for small values of $x$ (i.e.~when the slices are further away),
with the basic coincidence index being more strict and selective.    Also, the basic Jaccard 
tends to penalize these cases more intensely than the additive multised Jaccard index.

None of these indices are absolutely
better than the others.  It is the specific requirements of each application that should lead
to a suitable choice while considering the above identified properties of each index.
For instance, situations required enhanced selectivity and more strict similarity
quantification may consider the adoption of any of the two described coincidence indices.

\section{Continuous Densities and Scalar Fields} \label{sec:func}

The developments discussed in the previous sections
pave the way for considering also sets corresponding to densities, such as probability density
functions, as well as completely generic functions and scalar fields.   
One of the main problem to be overcome here is that densities often have infinite support, 
meaning that they extend over infinite ranges in their respective space.  

The problem of comparing two distributions is particularly important in many theoretical and
applied areas, having motivated great interest and the proposal of several respective approaches
(e.g.~\cite{Cha,multkolm}).

One interesting perspective that can be used to adapt the Jaccard and coincidence indices
so as to allow comparison of densities is developed as follows.  

\begin{figure}[h!]  
\begin{center}
   \includegraphics[width=0.9\linewidth]{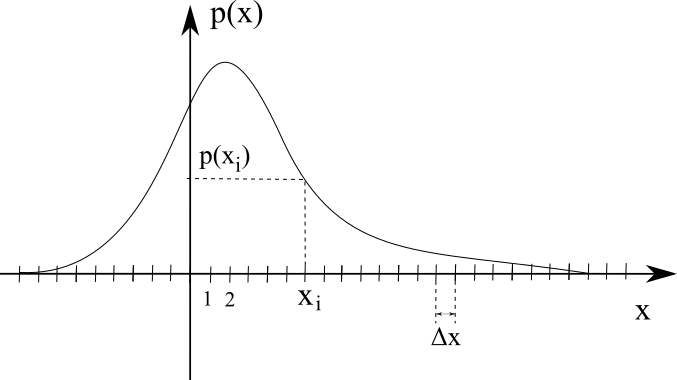}  
    \caption{A generic density function $p(x)$ being discretized with resolution $\Delta x$,
    so that it can be represented as a vector $\vec{p}=[x_i] \Delta x$. The integral of the original
    and respective discretization are assumed to be 1, so that they are both normalized as
    densities.}
    \label{fig:discrete}
    \end{center}
\end{figure}

We start by representing a generic continuous function in terms of a respective discretization,
with resolution $\Delta x$, as illustrated in Figure~\ref{fig:discrete}.

\vspace{0.5cm}

The density $p(x)$ becomes the vector $\vec{p} = [x_i] \Delta x$.  Now, in a vector the order
of the elements is all important, but it can indeed be incorporated into the multiset representation
as:
\begin{equation}
    X = \left\{ [x_1,m(x_1); \ldots;  [x_i,m(x_i)]; \ldots [x_n,m(x_n)] \right\}
\end{equation}

where $x_i$ are the elements in the respective support and 
$m(x_i)$ is the multiplicity of the element $x_i$ generalized to take real values.
In addition, we have also assumed, for simplicity's sake, that the discretization takes place
on $n$ points, which are henceforth understood as the \emph{support} of both the
function and the multiset.    The functions transformed into their respective multisets
are here called \emph{multifunctions}, or \emph{mfunctions}.

Now, let $m_A(x_i)$ and $m_B(x_i)$ correspond to the multiplicity of the elements $x_i$ in 
the two sets obtained by discretization of two density functions
$p_A(x)$ and $p_B(x)$ assuming the same support.  The respective
multiset Jaccard can now be obtained as:
\begin{equation}  \label{eq:Jac_P}
   \mathcal{J}_P(A,B) =   \frac{   \sum_{i \in \Phi} \min(m_A(x_i), m_B(x_i))  } 
    {\sum_{j \in \Phi} \max(m_A(x_j), m_B(x_j) )} 
\end{equation}
 
where $\Phi$ is the combined support of the two multisets $A$ and $B$.  

This index is then capable of expressing the similarity between the two original densities
up to the $x-$resolution $\Delta x$.  The above reasoning extends immediately to 
discrete probability densities.

By making $\Delta_x \rightarrow 0$, we then obtain:
\begin{equation}  \label{eq:Jac_P}
   \mathcal{J}_P(A,B) =   \frac{   \int_{\Phi} \min(m_A(x), m_B(x)) dx  } 
    {\int_{\Phi} \max(m_A(x), m_B(x)) dx } 
 \end{equation}
 
Observe that a bijective map is therefore obtained between the original density values
$\left[x, p(x) \right]$ and the respective 2-tuples $\left[x_i, (x_i)\right]$.

As such, the Jaccard index can be understood to correspond to a \emph{functional} derived
fro the two functions or, perhaps more specifically, an \emph{mfunctional}.

In addition to the above described limiting situation, it is also possible~\cite{CostaAnalogies} to consider
the Equation~\ref{eq:Jac_P} directly from the real function space, without the
need of taking the limit.  Observe that the generalized Jaccard index as defined by equation 
is completely specified and valid in the space of real functions, provided the integrals exist.

The above result extends immediately to density functions on higher dimensional domains as:
\begin{equation}  \label{eq:Jac_P}
   \mathcal{J}_P(A,B) =   \frac{   \int_{\Phi} \min(m_A(\vec{x}), m_B(\vec{x})) d\vec{x}  } 
    {\int_{\Phi} \max(m_A(\vec{x}), m_B(\vec{x})) d\vec{x} } 
 \end{equation}

Which provides a means to generalize the multiset Jaccard index to continuous or discrete
scalar or vector fields for any number of random variables.
 
\begin{figure}[h!]  
\begin{center}
   \includegraphics[width=1\linewidth]{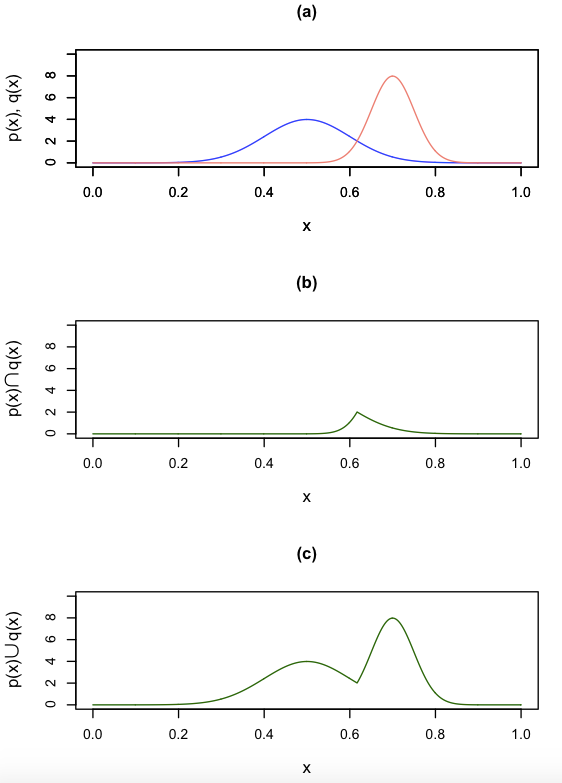}  
    \caption{Two probability density functions $p(x)$ and $q(x)$ (a), with
    respective intersection and union as shown in (b) and (c).  This situation yields a
    Jaccard index equal to 0.41.   The maximum value 1 is obtained whenever
    the two densities are identical.}
    \label{fig:ex_dens}
    \end{center}
\end{figure}

As an example of the generalization of multisets to real values, 
let's consider the two density  functions $X_A$ and $X_B$ depicted
in Figure~\ref{fig:ex_dens}(a).

\vspace{0.5cm}

The respective intersection $X_A \cap X_B$ and union $X_A \cup X_B$ 
of these two densities, obtained by using the minimum and maximum operation
between the elements of pair of values, are presented in 
Figures~\ref{fig:ex_dens}(a)  and (b), respectively.  The Jaccard index, obtained by
dividing the area of the intersection curve by the area of the union, yielded
the value 0.09257.

\begin{figure}[h!]  
\begin{center}
   \includegraphics[width=0.7\linewidth]{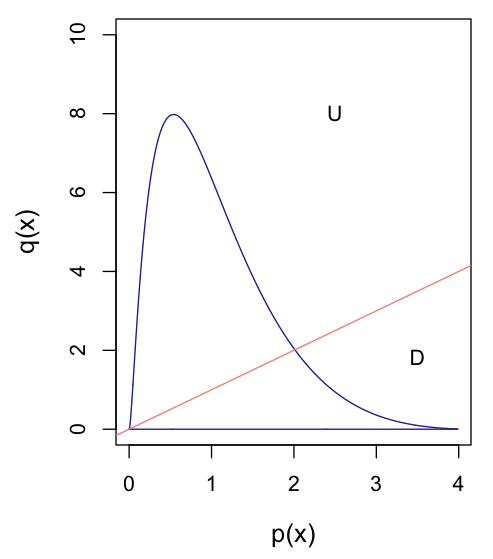}  
    \caption{The two probability densities $p(x)$ and $q(x)$ in Figure~\ref{fig:ex_dens}
    shown as a parametric curve in the respective scatterplot.  In case of discrete densities, they can be
    represented in terms of parametric curves related to the joint observations.
    Continuous densities can be represented in a similar manner.
    The identity line, shown in salmon, partitions
    the scatterplot space into the two regions $U$ and $D$. }
    \label{fig:parametric}
\end{center}
\end{figure}

It is also interesting to observe that the comparison of two densities can be represented as
a scatterplot, with the two density functions defining a parametric curve.  This is
illustrated in Figure~\ref{fig:parametric}.  The closest the set of points covered by the
respectively obtained parametric curve are to the identity line (in salmon), the higher
the similarity index between the two densities will be.  This construction can be immediately
extended to generic functions.

\vspace{0.5cm}

The Jaccard index can be also adapted to quantify the separation of two groups of points,
or clusters, which can be understood as a discrete or continuous scalar field.  
The basic idea here would be to represent each of the clusters in terms of the
joint probability density and then apply the Jaccard index over them by considering the
densities as the respective multiplicity of every element.  This method can be applied to
any number of involved features.

Though we have so far considered  both $X_A(\vec{x})$ and $X_B(\vec{x})$ to correspond to
non-negative scalar fields with hypervolume 1, it is actually possible to employ the
Jaccard and coincidence indices to quantify the similarity between any two scalar mfunctions
or mfields $\phi_A(\vec{x})$ and $\phi_B(\vec{x})$ sharing the same domain, even in
presence of negative multiplicities.  

The extension of similarity indices to negative values has been previously 
approached in~\cite{Mirkin,Akbas1,Akbas2}.  
A respective possible manner to adapt the Jaccard index to negative multiplicities in 
a two-dimensional  space is as follows.   In case the pair of points $[m(X_A),m(X_B)]$ 
is in the first quadrant, the minimum
and maximum between the two multiplicities are accumulated into the 
intersection and union integral, respectively.  It the point is in the
third quadrant, both coordinates have their signal inverted and the respective
minimum and maximum are accumulated.  

Otherwise, if the point belongs to the II or IV quadrant, the
point $[m(X_A),m(X_B)]$ is mirrored into the opposite quadrant respectively to the vertical axis,
and it is the negative of the minimum between the multiplicities that is then accumulated
into the intersection integral (to compensate for the
reflection), while the union is taken with positive sign, and the resulting accumulated
intersection and union are then used in Equation~\ref{eq:Jac_P} to obtain the respective
Jaccard index.

\begin{figure}[h!]  
\begin{center}
   \includegraphics[width=0.7\linewidth]{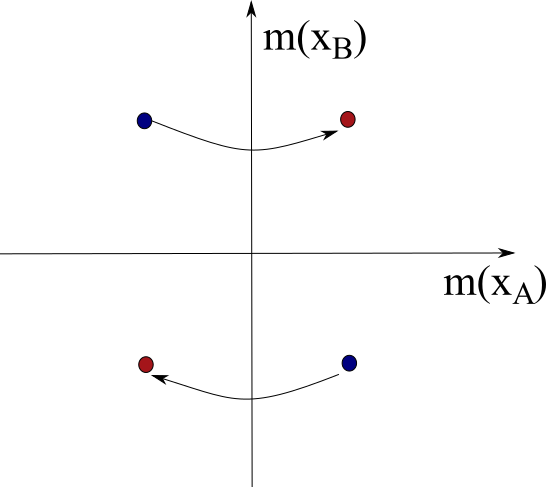}   
    \caption{Jaccard for mfunctions with negative multiplicity.
    Points in the II and IV quadrants are reflected with respect to the vertical
    axes, and their intersection (minimum values) and union (maximum values) enter
    with negative values in the accumulated intersection and union.}
    \label{fig:cos_sin}
    \end{center}
\end{figure}
\vspace{0.5cm}

For instance, let's calculate the multisets Jaccard index for the functions $f(x) = cos(\theta)$
and $g(x) = sin(\theta)$ for a complete period $0 \leq \theta \leq 2 \pi$, as illustrated
in Figure~\ref{fig:cos_sin}.

\begin{figure}[h!]  
\begin{center}
   \includegraphics[width=0.8\linewidth]{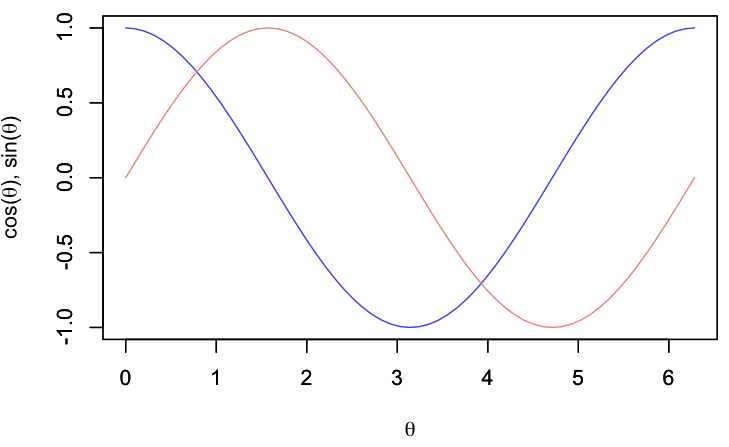}   \\ (a) \\ \vspace{0.3cm}
   \includegraphics[width=0.8\linewidth]{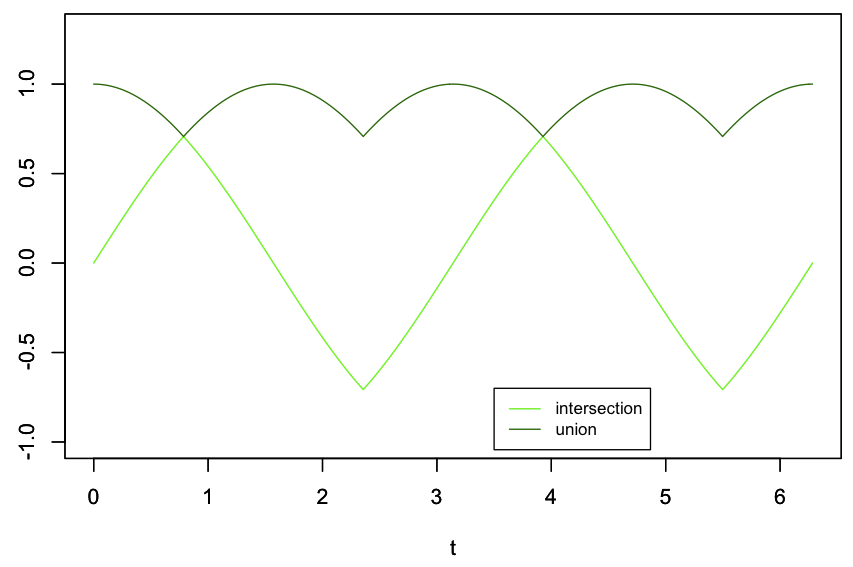}   \\ (b) \\ 
    \caption{The Jaccard index calculated for a cosine and a sine function.
    The mfunctions are shown in (a), and the respective interesection and union
    mfunctions are shown in (b).  The obtained Jaccard index was equal to 0.
    Indices of 1 and -1 will be obtained in case $g(x) = cos(t)$ and $g(x) = -cos(t)$,
    respectively.}
    \label{fig:cos_sin}
    \end{center}
\end{figure}
\vspace{0.5cm}

This operation, which has been found to correspond to the intersection of real multisets
with possibly negative values~\cite{CostaMset,CostaGenMops}, can be summarized as:
\begin{equation}
    f(x) \sqcap g(x) = \int_{S}  s_f s_g \min(s_f f(x), s_g g(x)) dx 
\end{equation}

where $S$ is the combined support of $f(x)$ and $g(x)$.

From which, the multiset convolution~\cite{CostaMset}  (\emph{mconvolution})
of two functions can be derived:
\begin{equation}  \label{eq:jac_Cont}
  f(x) \Box g(x) [y] = \int_{S}  \frac{  f(x) \sqcap g(x-y) } {f(x) \tilde{\sqcup} g(x-y)} dx
\end{equation}

where $f(x) \tilde{\sqcup} g(x-y)$ corresponds to the absolute value union of generalized
multisets~\cite{CostaGenMops}:
\begin{equation}
    f(x) \tilde{\sqcup} g(x) = \int_{S}  s_f s_g \min(s_f f(x), s_g g(x)) dx 
\end{equation}

Equation~\ref{eq:jac_Cont} presents a direct analogy to the multiset Jaccard index.
The other generalizations of the Jaccard index can be readily employed in the
above expressions in order to cater for less or more strict similarity quantification.

Preliminary results have shown that the multiset convolution provides, in general,
sharper and more selective peaks and smaller sidelobes than the standard 
correlation~\cite{CostaMset,CostaComparing}.

\begin{figure}[h!]  
\begin{center}
   (a) \\
   \includegraphics[width=0.7\linewidth]{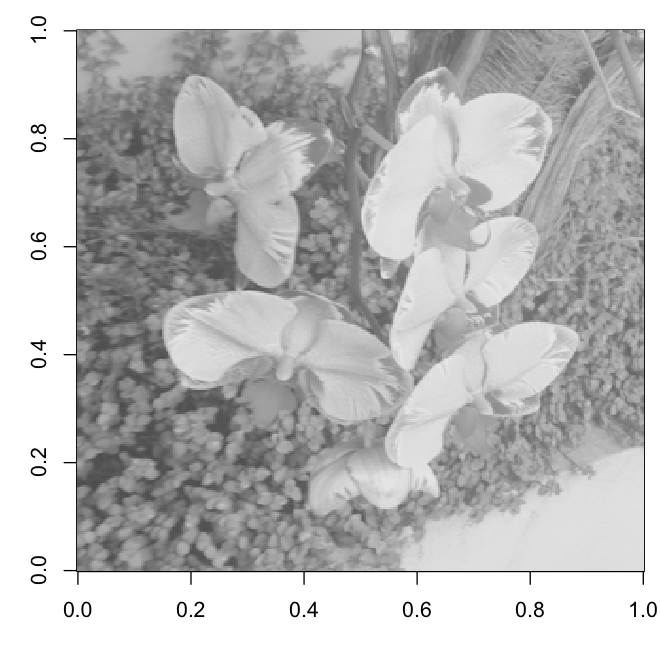}  \\  \vspace{0.2cm} (b) \\
   \includegraphics[width=0.7\linewidth]{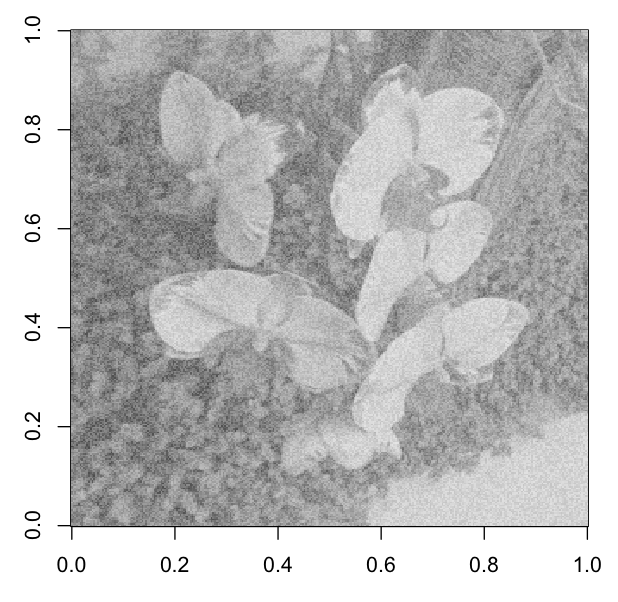}  \\  \vspace{0.2cm} (c) \\
   \includegraphics[width=0.7\linewidth]{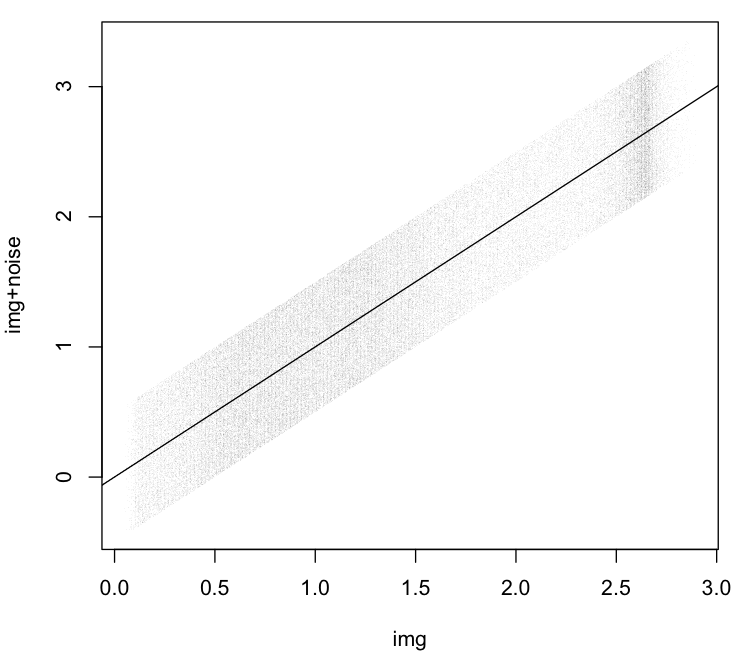}  
    \caption{A gray level image of flowers $img[x,y]$ (a) was mixed with random noise
    uniformly distributed between $-0.5$ and $0.5$, resulting in the noisy image
    $img[x,y] + \xi[x,y]$ shown in (b). The resulting scatterplot is depicted in (c), 
    including the identity line defining the two regions for calculation of the
    scalar field intersection and union, from which 
    a respective Jaccard index of $\mathcal{J}(img,img+\xi) = 0.83$ was obtained,
    reflecting a relatively high similarity between the two scalar fields. }
    \label{fig:flower}
    \end{center}
\end{figure}

A further example of the Jaccard index adapted to multidimensional scalar fields,
namely a gray level image, also incorporating the respective scatterplot representation
of the paired multiplicities is provided in Figure~\ref{fig:flower}.

\vspace{0.5cm}

\section{Joint Variations}

Joint variation are often taken in a normalized manner as when using the Pearson correlation coefficient.
More specifically, we have that this coefficient can be understood as corresponding to the variance provided
the samples of the two sets have been first standardized.  By standardization it is henceforth
understood that, given a random variable $X$, we apply the following random variable
transformation:
\begin{equation}
   \tilde{X} = \frac{X - \mu_X} {\sigma_X}
\end{equation}

This standardization has the effect of normalizing the dispersions of a random variables, so that
the its variance becomes 1 while the average is 0.  It can also be verified that a standardized
random variable will present most of its observations within the interval $[-2,2]$.

In the case of a set of $N$ observations of two standardized random variables, the
Pearson correlation coefficient becomes:
\begin{equation}
   \mathcal{P}(X,Y) =  \frac{1}{N} \sum_{i,j=1}^{N}  {[\tilde{X}_i][\tilde{Y}_i]} 
\end{equation}

When two standardized random variables  $\tilde{X}$ and $\tilde{Y}$ are taken jointly, they 
define a \emph{scatterplot} providing a useful illustration about the interrelationship between the
two considered values.   This scatterplot can be immediately understood as corresponding
to a sampling of the joint probability density of the two random variables, which may be kernel expanded
to obtain an estimation of the respective counterpart.

The quantification of joint variation by a L1-based operatoron possibly 
negative values~\cite{Akbas2}  has been previously addressed~\cite{CetinPCA}.

\begin{figure*}[h!]  
\begin{center}
   \includegraphics[width=1\linewidth]{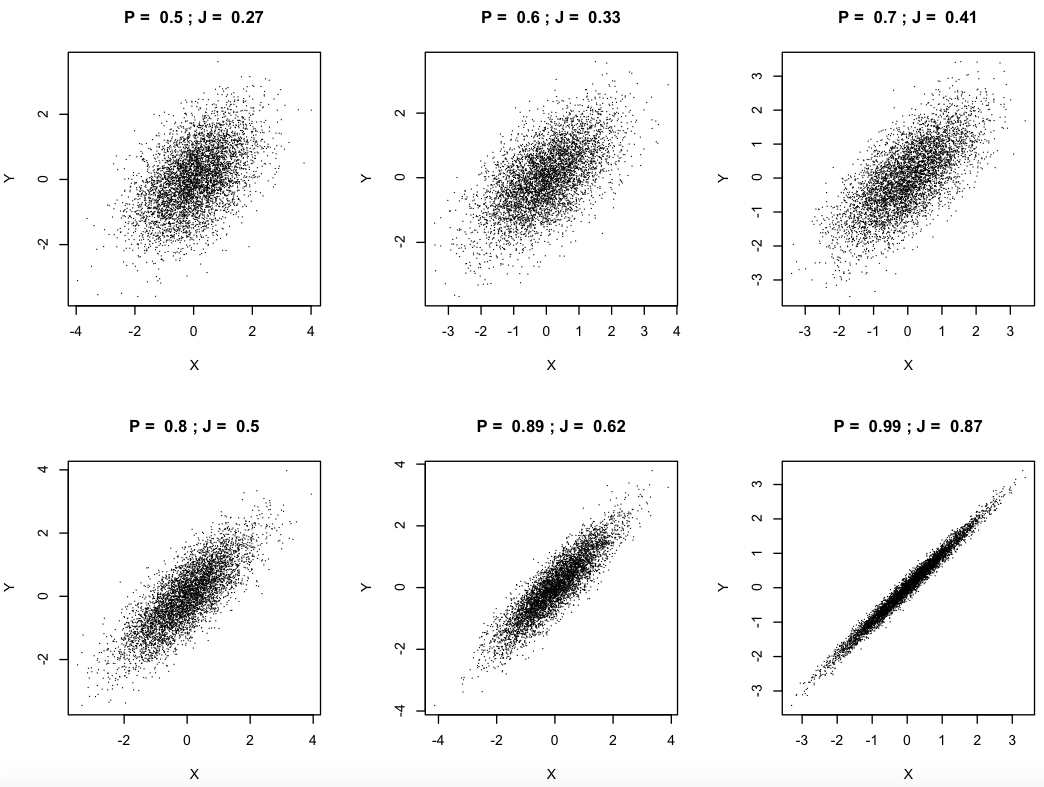}  
    \caption{Comparision of the Pearson correlation coefficient and the multiset Jaccard
    index for negative multiplicities with respect to several distribution of points with
    increasing correlation.  Interestingly, the Jaccard index seems to provide a more
    gradual quantification of the joint variations that is probably more compatible with
    our perception.  At the same time, the respective Pearson correlation coefficients tend to
    saturate as the correlation increases.}
    \label{fig:gaussians}
    \end{center}
\end{figure*}

It constitutes an interesting issue to consider joint variation quantification based 
on the Jaccard and coincidence indices.
In order to illustrate the possibility to quantify the joint variation of observations
in a scallterplot (or, actually, joint densities), we consider the situation in Figure~\ref{fig:gaussians},
which shows several scatterplots drawn from normal densities with increasing correlation.

\vspace{0.5cm}

\section{Multiple Sets} \label{sec:multi}

We have so far considered indices applied to two sets or entities.  There are two basic ways in which
more sets can be taken into account.  The first one consists of simply understanding that each of the
two sets $A$ and $B$ are obtained by set operation combinations among several other sets.

For instance, in case we are interested in $A = (C \cap D) \cup E - F$ and $B = C \cup G$, we
can  write:
\begin{eqnarray}
   A = f(C, D, E, F)  \nonumber \\
   B = f(C,G)  \nonumber
\end{eqnarray}

and then apply the Jaccard or coincidence indices.
   
Observe that there is absolute no restriction on these functions, except that they are not
both empty sets.

The Jaccard index for the example above can be expressed as:
\begin{equation}  \label{eq:Jac}
   \mathcal{J}(A(C,D,E,F),B(C,G)) =   \frac{\left| A(C,D,E,F) \cap B(C,G) \right|}  {\left| A(C,D,E,F) 
   \cup B(C,G) \right|}  \nonumber
\end{equation}

Therefore, a vast range of possible combinations of diverse sets become possible, but they will
ultimately always lead to two resulting sets $A$ and $B$ to be compared by the Jaccard or
coincidence indices.

There is another interesting possibility to take into account more than 2 sets, and this corresponds
to extending the Jaccard index, for instance in the case involving 3 sets, as:
\begin{equation}  \label{eq:Jac}
   \mathcal{J}_3(A,B,C) =   \frac{\left| A \cap B \cap C \right|}  {\left| A \cup B \cup C\right|} \nonumber
\end{equation}

with $0 \leq \mathcal{J}(A,B,C) \leq 1$.  This concept can be immediately 
extended to any number $N_S$ of sets.

The extension of the interiority index becomes:
\begin{equation} \label{eq:Int}
   \mathcal{I}_{[3,1]}(A,B,C) =   \frac{\left| A \cap B \cap C \right|}  {\min \left\{ \left| A \right|, 
   \left| B \right|,\left| C \right| \right\} }  \nonumber
\end{equation}

It can be verified that this extended interiority index now quantifies how much the smallest of the 
sets is contained in the overall intersection.   However, it does not take into account how the
intermediate size set relates to the mutual intersection.  This can be accomplished by introducing
a second interiority index as:
\begin{equation} \label{eq:Int}
   \mathcal{I}_{[3,2]}(A,B,C) =   \frac{\left| A \cap B \cap C \right|} 
    { \left| X \right| }  \nonumber
\end{equation}

where $X$ is the set with the second smallest cardinality.

The two obtained interiority indices can then be combined into a single respective index as:
\begin{equation} \label{eq:Int}
   \mathcal{I}_{3}(A,B,C) =      \mathcal{I}_{[3,1]}(A,B,C)  \;  \mathcal{I}_{[3,2]}(A,B,C)      \nonumber
\end{equation}

with $0 \leq \mathcal{I}_{3}(A,B,C) \leq 1$.

We can now define the coincidence index extended to three sets as:
\begin{equation} \label{eq:Int}
   \mathcal{C}_{3}(A,B,C) =      \mathcal{I}_{3}(A,B,C)  \;  \mathcal{J}_{3}(A,B,C)      \nonumber
\end{equation}

A similar development applies to more than 3 sets.

\begin{figure}[h!]  
\begin{center}
   \includegraphics[width=0.7\linewidth]{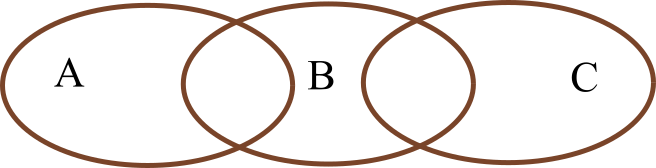}  
    \caption{Three sets $A$, $B$ and $C$ characterized by sequential, or chained
    intersections. In the suggested approach, $B$ is taken as a candidate reference
    for intermediating the other two sets through a chaining relationship.}
    \label{fig:chaining}
    \end{center}
\end{figure}

The consideration of more than 2 sets in similarity index suggests other possible extensions of
the Jaccard and coincidence indices.  For instance, it becomes interesting not only to quantify
the overall similarity between 3 sets, but also to develop indices capable of reflecting how these
three sets are connected one another.  Consider the situation depicted in Figure~\ref{fig:chaining}.

\vspace{0.5cm}

This situation suggests that set $B$ intermediates the connection between
the sets $C$ follows and $A$, therefore establishing a \emph{chaining} relationship.   
The Jaccard index with 2 sets cannot cope directly with this situation.

A possible index involving three sets that can quantify the chaining between 3 sets is:
\begin{equation}
  \mathcal{X}(A,B,C) = \mathcal{J}(B,(A \cap B) \cup (B \cap C)) \; \left[ 1 - \mathcal{J}(A,C) \right] \nonumber
\end{equation}

As an example, let's consider:
\begin{eqnarray}
 A = \left\{ a, b, c, d, e, f, g \right\};  \nonumber \\
 B = \left\{ e, f, g, h, i, j, k \right\}; \nonumber \\
 C = \left\{ i, j, k, l, m, n, o \right\} \nonumber
\end{eqnarray}

It folows that:
\begin{eqnarray}
 A \cap B = \left\{ e, f, g \right\}; \nonumber \\
 A \cap C = \left\{  \right\}; \nonumber \\
 B \cap C = \left\{ i, j, k \right\}; \nonumber \\
 A \cup C = \left\{ a, b, c, d, e, f, g, i, j, k, l, m, n, o \right\}; \nonumber \\
 (A \cap B) \cup (B \cap C) = \left\{ e, f, g, i, j, k\right\};  \nonumber \\ 
 B \cap [(A \cap B) \cup (B \cap C)] = \left\{ e, f, g, i, j, k\right\}  \nonumber \\
 B \cup [(A \cap B) \cup (B \cap C)] = \left\{ e, f, g, h, i, j, k\right\}  \nonumber 
\end{eqnarray}
 
So, we have that: 
\begin{eqnarray}
   \mathcal{J}(B,(A \cap B) \cup (B \cap C)) =  \nonumber \\
   = \frac{\left|  B \cap [(A \cap B) \cup (B \cap C)]  \right|} 
    {\left|  B \cup [(A \cap B) \cup (B \cap C)]  \right|} = \frac{6}{7}
\end{eqnarray}

and:
\begin{eqnarray}
   \mathcal{J}(B,C) =  \frac{|A \cap C|} {|A \cup C|} = \frac{0}{14} = 0
\end{eqnarray}

From which we obtain the chaining index value of:
\begin{eqnarray}
   \mathcal{X}(A,B,C) = \nonumber \\
   =\mathcal{J}(B,(A \cap B) \cup (B \cap C)) \; \left[ 1 - \mathcal{J}(A,C) \right] = \nonumber \\
   =\frac{6}{7} \left[1 - 0  \right] =  \frac{6}{7}  \nonumber
\end{eqnarray}

which provides an interesting indication of the chaining between the sets $A$, $B$, and $C$.
Observe that the above described approach assumes that set $B$ has been adopted as
a reference for implementing the chaining between $A$ and $C$.  More generic situations
can be addressed by considering successive pairwise combinations.

It should be observe that it is possible that one of the intersections betwen $B$ and $A$ or
$C$ is large enough to bias the above index.  In these situations, it is possible to incorporate
an additional index specifying a minimum overlap between both $A$ and $B$ as well as
$B$ and $C$.

Several other analogous chaining indices involving 3 or more sets or other structures
are possible, leading to complementary properties.

\section{The Jaccard and Coincidence Indices in Modeling} \label{sec:modeling}

By allowing several types of mathematical structures to have their relationships
to be quantified in terms of respective indices, it becomes possible to objective and
quantitatively address a wide range of theoretical and practical problems, while also
catering for the consideration of stochasticity.

In addition, the several indices
discussed and suggested in this work represent a valuable resource while developing
models (e.g.~\cite{CostaModeling}) through the combination of datasets as described
in~\cite{CostaAmple}.

Then, we have several additional possibilities of applying these indices.  For instance,
a new dataset can be compared to those already modeled by using the similarity
indices.  Also of particular interest is to identify which combinations, through set
operations, between the existing datasets associated to models are more likely
to account for other datasets of interest, therefore providing insights about how
respective models can be identified, related, or developed.

The discussed indices are also interesting from the perspective of characterizing,
developing, validating and applying pattern recognition and deep learning 
approaches~\cite{HInton,Schmidhuber,CostaDeep}.

\section{Concluding Remarks}

Relationships between the several important mathematical structures --- including 
sets, functions, vectors, densities, and graphs --- are critically important in virtually
all areas where mathematics is employed.  Given its interesting features, the Jaccard
index has been extensively employed in a large range of scientific and technological
situations.  Also as a consequence of its potential, the Jaccard index has been
generalized in several manners.

The present work aimed at generalizing further the Jaccard index.  One of the first
discussed possibilities consisted in using the interiority index, capable of quantifying
how much a set is contained into another, as means to complement a identified
limitation of the Jaccard index in taking into account the interiority of one set into
the other .  This index was then combined with the
Jaccard index to yield the \emph{coincidence} index, which is believed to provide
a more strict and selective quantification of the similarity between sets.  The possibility to
adopt the sum of multisets instead of the union was also addressed, with promising
results for the situations where the multiplicity of the elements have to be fully
taken into in account.

The possibility to apply the Jaccard and coincidence indices on continuous sets
was then addressed by considering the areas of the involved regions in place of the
number of set elements.   This adaptation of the Jaccard index
allowed the consideration of density fields and functions, which was approached by
using the Jaccard index for multisets.    The potential of this generalization of the
Jaccard index was then briefly illustrated with respect to probability density functions
as well as in a comparison between the cosine and sine functions, which are
not normalized and can take negative values, as well as a real-world image and
a respective noise version.

The intrinsic relationship between similarity indices and statistical quantifications of
joint variation between random variables was approached subsequently, and it
has been argued that both the Pearson correlation coefficient can be used to compare
two density functions, but also that a respective adaptation of the Jaccard and
coincidence indices can also be used for that finality.  We also discussed the
interesting possibility to visualize the action of the Jaccard and coincidence indices
with respect to the division of the data into two regions defined by the
identity line in the scatterplot distribution.  

The also interesting situation of similarity and other indices considering three or more
sets was then discussed, identifying the possibility to consider the two sets involved
in the basic Jaccard and coincidence index as corresponding to the result of set operation
combinations between any number of other sets.  Another important extension was
considered with respect to taking into account more than 2 sets as arguments for
the similarity indices, which was illustrated in terms of a suggested index to quantify
the chaining between three sets.

Several are the further possible works motivated by the concepts and methods
reported and suggested in this work, a more complete list of which would be particularly
extensive.  Some of the possibilities include comparing the described indices with other
indicators of similarity, the identification of other types of relationships that can be 
quantified when considering 3 or more sets and analogue generalizations of other
interesting indices, as well as extending the described indices to
other mathematical structures.  

In addition, as observed in Section~\ref{sec:modeling},
similarity and other indices such as those addressed here provide valuable means for
developing and evaluating \emph{models} of data as well as for several pattern recognition
and deep learning tasks.

\vspace{0.7cm}
\emph{Acknowledgments.}

Luciano da F. Costa
thanks CNPq (grant no.~307085/2018-0) and FAPESP (grant 15/22308-2).  
\vspace{1cm}

\bibliography{mybib}
\bibliographystyle{unsrt}

\end{document}